\documentclass[10pt,twocolumn,letterpaper]{article}

\usepackage{cvpr}
\usepackage{times}
\usepackage{epsfig}
\usepackage{graphicx}
\usepackage{amsmath}
\usepackage{amssymb}

\usepackage{color}
\usepackage{amsthm}
\usepackage{dblfloatfix}
\usepackage{tabls}
\usepackage{caption,float}
\usepackage{dblfloatfix}
\usepackage{url}
\usepackage[ruled,vlined]{algorithm2e}
\usepackage[export]{adjustbox}
\usepackage{soul}
\usepackage{multirow}

\usepackage{algpseudocode}

\DeclareMathOperator*{\argmin}{arg\,min}

\definecolor{myPurple}{rgb}{0.4, .0, .8}
\definecolor{myGreen}{rgb}{0, .8, .3}
\definecolor{myRed}{rgb}{0.8, .2, .2}
\definecolor{myOrange}{rgb}{0.8, 0.45, 0.0}
\definecolor{myBlue}{rgb}{.0, .0, 1.0}

\newcommand{\todo}[1]{\textcolor[rgb]{1.0,0.,0.}{[#1]}}

\usepackage[pagebackref=true,breaklinks=true,letterpaper=true,colorlinks,bookmarks=false]{hyperref}

\cvprfinalcopy 

\def\cvprPaperID{4998} 

\ifcvprfinal\fi
\begin{document}

\title{Self-Supervised Human Depth Estimation from Monocular Videos}

\author{Feitong Tan$^{1,*}$ \ \ Hao Zhu$^{2,}$\thanks{These authors contributed equally to this work.} \ \ Zhaopeng Cui$^{3}$ \ \ Siyu Zhu$^{4}$ \ \ Marc Pollefeys$^{3}$ \ \ Ping Tan$^{1}$ \\
$^{1}$ Simon Fraser University \ \ $^{2}$ Nanjing University  \\ $^{3}$ ETH Z\"urich  \ \ $^{4}$ Alibaba AI Labs  \\
}

\maketitle

\begin{abstract}
   Previous methods on estimating detailed human depth often require supervised training with `ground truth' depth data.
   This paper presents a self-supervised method that can be trained on YouTube videos without known depth, which makes training data collection simple and improves the generalization of the learned network.
   The self-supervised learning is achieved by minimizing a photo-consistency loss, which is evaluated between a video frame and its neighboring frames warped according to the estimated depth and the 3D non-rigid motion of the human body.
   To solve this non-rigid motion, we first estimate a rough SMPL model at each video frame and compute the non-rigid body motion accordingly, which enables self-supervised learning on estimating the shape details. 
   Experiments demonstrate that our method enjoys better generalization and performs much better on data in the wild.

\end{abstract}
\vspace{-0.1in}

\section{Introduction}

Understanding and reconstructing human motion from images and videos is an important problem in computer vision with many applications including surveillance, VR/AR,  and tele-presence. Many works focus on estimating a 2D or 3D skeleton model~\cite{cao2017realtime, newell2016stacked, Martinez_2017_ICCV, pavlakos2017coarse, mehta2017vnect, sun2018integral}. While a skeleton model could be useful for surveillance, other applications demand a 3D surface model of the undressed human body, which is often represented by the SMPL~\cite{SMPL:2015} or SCAPE~\cite{anguelov2005scape} model. Many works have been proposed to estimate those parametric shape models from images~\cite{pavlakos2018learning, kanazawa2018end, omran2018neural, guler2019holopose, kanazawa2019learning}. 
However, the mid- and high-frequency shape details,~\eg, cloth wrinkles and folds, are not captured in the SMPL and SCAPE models, which limits their application in AR/VR and tele-presence applications. Only a handful of recent works \cite{tang2019neural,zhu2019detailed,saito2019pifu,zheng2019deephuman,alldieck2019tex2shape, Bhatnagar_2019_ICCV} can recover those details from a single image, but they all rely on ground-truth 3D data for supervision.
This paper aims to develop a self-supervised method for detailed human depth estimation, such that the network can be trained on a much larger dataset, e.g. YouTube videos, for improved performance. 

Self-supervised learning has been adopted \cite{godard2017unsupervised, garg2016unsupervised, kuznietsov2017semi} to train depth estimation from a single image for static scenes by enforcing photo-consistency between the left and right views in a stereo pair. Basically, the left view can be warped to the right view according to the estimated depth, the photo-consistency between the right view and the warped left view can be used to train the network. 
In principle, if the human motion between two video frames is known, we could adopt the same photo-consistency principle to train the network for human depth estimation. However, the challenge is that human body has non-rigid motion and requires much more complicated motion models such as those in DynamicFusion~\cite{newcombe2015dynamicfusion} and VolumeDeform~\cite{innmann2016volumedeform}, which are difficult to estimate as well. 




To address this challenge, we represent the human depth by a SMPL model with an additive residual detail map. This representation is similar to the base  and detail shape formulation in ~\cite{tang2019neural}, but it bears two important advantages. 
Firstly, the SMPL model estimation has been well studied and is robust even on data in the wild, which makes the base shape estimation more reliable. As we will see in experiments, this helps to reduce large errors in the estimated depth.
Secondly, the SMPL model parameters have clear semantic meanings and can be used to induce the non-rigid motion of the human body between two neighboring video frames.
In this way, the non-rigid human body motion can be solved, and the photo-consistency for self-supervised learning in \cite{godard2017unsupervised, garg2016unsupervised, kuznietsov2017semi} can be employed to train the depth estimation network.

Measuring photo-consistency between neighboring video frames is still hard, even when the non-rigid motion to align the two human shapes is known. We design our method to be robust to occlusion, motion inaccuracy, and shading changes to achieve a robust method.



With the proposed self-supervised framework, we can train our network using almost endless online video clips. This vast training data significantly improves the generalization of the trained network on unseen data, making human depth estimation more robust.

\section{Related Work}


\textbf{Skeleton Pose or Parametric Model Estimation.}  With the development of deep neural network, the estimation of 2D skeleton joints~\cite{cao2017realtime, newell2016stacked} and 3D skeleton joints~\cite{Martinez_2017_ICCV, pavlakos2017coarse, mehta2017vnect, sun2018integral} has achieved great success with robust performance. 
Many other works focus on estimating an undressed human body shape from a single image, as the skeleton joints are insufficient to convey shape information. The undressed body shape is often represented by the SCAPE~\cite{anguelov2005scape}, SMPL~\cite{SMPL:2015}, or SMPL-X~\cite{pavlakos2019expressive} model, which encodes the body shape by the pose and shape parameters. These models can be fitted according to estimated skeleton joints~\cite{bogo2016keep, lassner2017unite} or be directly regressed as in~\cite{guan2009estimating, dibra2016hs-nets, tan2017indirect, pavlakos2018learning, kanazawa2018end, omran2018neural,pavlakos2019texturepose, guler2019holopose,tung2017self, kanazawa2019learning}.

While the estimation of skeleton pose and these parametric body shapes are relatively well studied and robust, they are insufficient for certain applications such as tele-presence. In comparison, we strive to recover detailed human shapes, which has broader applications.


\textbf{Non-parametric shape estimation.}  
As the parametric SMPL or SCAPE captures limited shape details, non-parametric representations have been adopted in human shape estimation. Varol~\etal~\cite{varol2018bodynet} and Venkat~\etal~\cite{venkat2018deep} used a 3D volumetric model to represent human shapes for better flexibility.
G{\"u}ler~\etal\cite{alp2018densepose} recovered a dense 2D-to-3D surface correspondence field for the human body, and the SMPL model can be generated according to the correspondence.
Zhu~\etal~\cite{zhu2018view} and Rematas~\etal~\cite{rematasCVPR18} directly predicted the depth map from a single image by training on synthetic data. Li~\etal~\cite{li2019learning} exploited motion  parallax  cues  from  static  scenes  to  guide the  human depth  prediction by watching 'frozen people'.
Kolotouros~\etal~\cite{kolotouros2019convolutional} directly regressed the vertices in the SMPL model while retaining the topology.  
Natsume\etal~\cite{natsume2019siclope} used 2D silhouettes and 3D joints of a body pose to describe the immense human shape.

All the above methods still cannot capture shape details such as cloth wrinkles. Only a handful of recent works\cite{tang2019neural,zhu2019detailed,saito2019pifu,alldieck2019tex2shape,zheng2019deephuman, Bhatnagar_2019_ICCV} are able to recover those details. Among them,
Tang~\etal~\cite{tang2019neural} proposed a base + detail shape representation, Zheng~\etal~\cite{zheng2019deephuman} followed the volumetric shape representation, Zhu~\etal~\cite{zhu2019detailed} and Alldieck~\etal~\cite{alldieck2019tex2shape} improved the undressed SMPL model with hierarchical morphable models and vertex displacements respectively. 
Saito~\etal~\cite{saito2019pifu} defined a pixel-aligned implicit function to represent the human shape.
Bhatnagar~\cite{Bhatnagar_2019_ICCV} designed a neural network to estimate the garment geometry separately, and then dress the SMPL model with the garments. 

However, all of them require ground-truth 3D data for supervised training, which is hard to obtain and could lead to serious generalization problem.
We also aim to recover human shape details. But we advocate for self-supervised learning to exploit the vast online videos for training. Our approach significantly improves the network performance on  in-the-wild data.
\textbf{Self-supervised Depth Estimation.} To alleviate the demands on the expensive ground-truth 3D data, various self-supervised approaches have been proposed to train depth prediction networks. These methods typically train the network by minimizing a photo-consistency loss for some view synthesized according to the inferred depth.
The methods in~\cite{godard2017unsupervised, garg2016unsupervised, kuznietsov2017semi} utilize stereo images with known relative motion between the left and right views. 
Some methods~\cite{xie2016deep3d, zhou2017unsupervised,wang2018learning, yin2018geonet} 
use a monocular moving camera and enforce photo-consistency between neighboring video frames. This setting is more challenging, since these methods need to estimate the camera motion at the same time of estimating scene depth.
Khot~\etal~\cite{khot2019learning} designed a self-supervised method for multi-view stereo with a sophisticated loss function dealing with occlusion and shading changes.

Our method also takes the self-supervised approach to train a depth estimation network. But our method is designed for a moving human, instead of the static scene in all the above methods. Thus, the motion model between our neighboring video frames are much more complicated than those works.

\section{Methods}\label{sec:method}
\vspace{-0.1in}
\begin{figure*}[t]
\begin{center}
\includegraphics[width=0.9\linewidth]{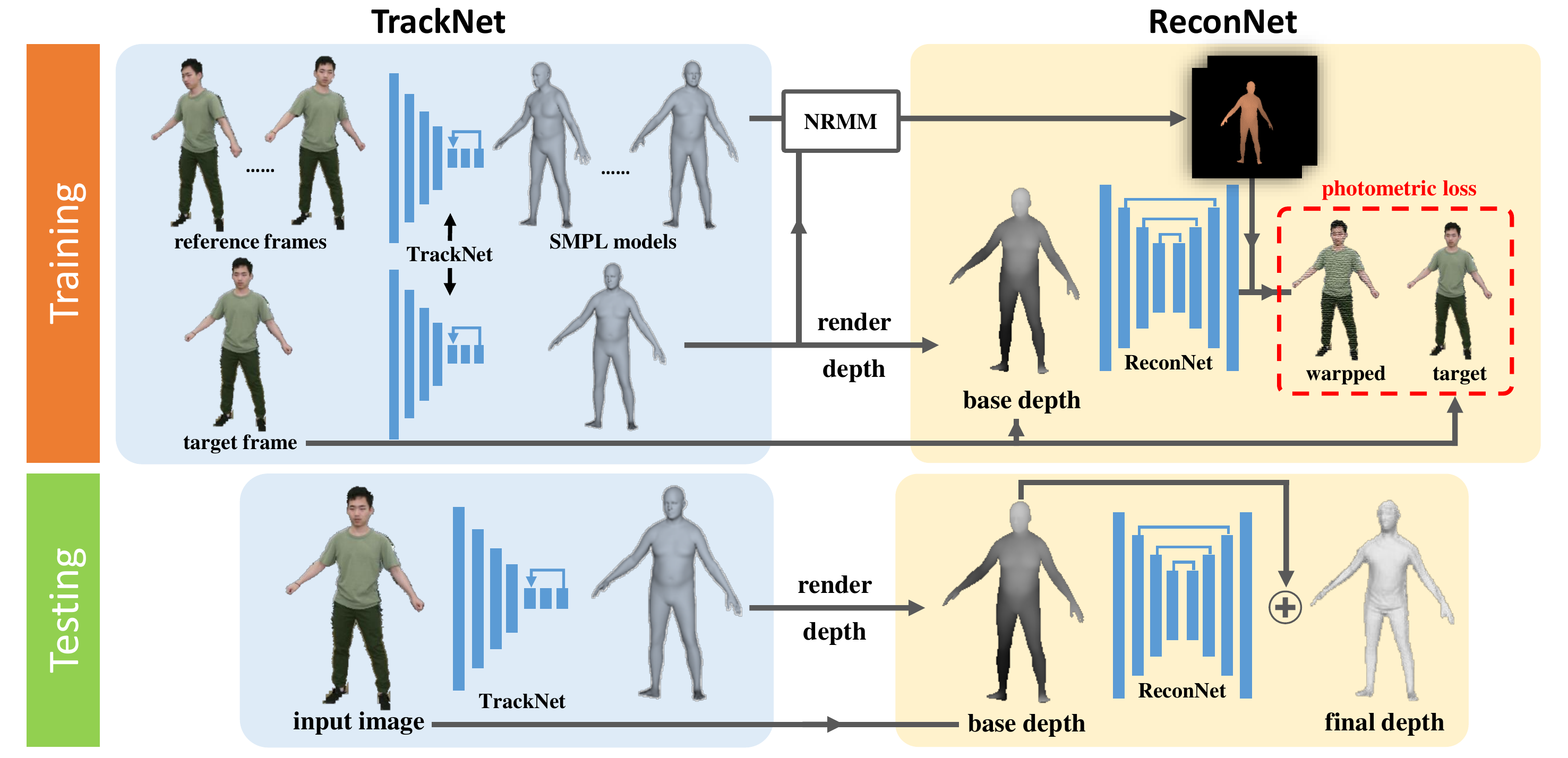}
\end{center}
    \vspace{-0.25in}
    \caption{Overview of our system. At training time, our system includes the TrackNet to compute a SMPL model at each input video frame, the NRMM model to compute the non-rigid motion to align two neighboring human shapes, and the ReconNet to estimate shape details. At testing time, our system first compute the SMPL model from the TrackNet and then estimates the details form the ReconNet and  combine them as the final result. }
    \vspace{-0.15in}
\label{fig:pipeline}
\end{figure*}


An overview of the proposed framework is shown in Figure~\ref{fig:pipeline}. It composes of three main components: (1) TrackNet, a neural network to estimate a Skinned Multi-Person Linear (SMPL) model~\cite{SMPL:2015}  from a single image, which determines the base depth and also the non-rigid motion between consecutive frames. (2) NRMM, 
a module to compute the non-rigid motion according to the SMPL model to align the 3D human shapes at neighboring frames. (3) ReconNet, a neural network trained in a self-supervised manner to estimate the residual detail shape, which will be added to the base depth to capture shape details. 


In the training stage, a short video sequence of a person is fed to the TrackNet to compute a SMPL model for each frame. In the next, a target frame is selected, and the non-rigid motions from the other frames to the target frame are computed through the NRMM module according to their SMPL models. Finally, the ReconNet will be trained in an self-supervised manner given the non-rigid motion by enforcing the photo-consistency loss. 
In the prediction stage, the TrackNet will estimate the SMPL model to generate the base depth, with which the ReconNet will predict the details. The final result is simply the addition of the base shape and the details.

\subsection{Pose Tracking}\label{sec:pose_track}
We estimate a SMPL model at each frame to capture the human pose and rough shape. SMPL is a parameteric undressed human body model with 72 pose parameters $\theta$ and 10 shape parameters $\beta$ controlling a triangle mesh of 6,890 vertices. The parameters $\theta$ define the 3D rotation of each skeleton joint, and the parameters in $\beta$ describe the height, weight, and other body shape metrics. Compared with a skeleton model, a SMPL model encodes strong human shape prior and provides more information including limbs orientation and the rough shape.

We design a TrackNet module to predict SMPL parameters for each frame of the input video. The undressed body shape defined by SMPL parameters are used as the base shape for further process. 
Compared to estimating an non-parametric base shape in~\cite{tang2019neural}, our approach is more robust since the base shape is constrained into a much smaller parameter space with strong human shape priors.

Our TrackNet adopts the same network architecture as HMR~\cite{kanazawa2018end}, which consists of a ResNet-50~\cite{he2016identity} as a feature extractor and an iterative error feedback regressor~\cite{carreira2016human}.  The original HMR model is trained on images with annotated 2D joints.
In order to produce more accurate results, we captured a small set of videos with `ground-truth' SMPL coefficients generated by DoubleFusion~\cite{yu2018doublefusion} and further finetuned the TrackNet after pretraining following \cite{kanazawa2018end}. 

The TrackNet outputs a 85-D vector, with 82 parameters as SMPL coefficients and 3 parameters for the weak-perspective camera model. The loss function to finetune TrackNet on our DoubleFusion data is formulated as:
\begin{equation}
L_{tn} =   L_{para} + \theta_{p}L_{J\_pos} + \theta_{r}L_{J\_rot}
\label{equ:loss_tn}
\end{equation}
where $L_{para}$ is the $L_1$ loss of SMPL parameters, $L_{J\_pos}$ and $L_{J\_rot}$ are the $L_1$ loss of 3D position and rotation of SMPL joints respectively.  $\theta_{p}$ and $\theta_{r}$ are the loss weights, and both of them are set to $1$ in our experiments. 
Please note that our method is not limited to the specific HMR~\cite{kanazawa2018end} model and  the TrackNet can be upgraded with other SMPL model estimation networks. 

\subsubsection{Camera Model Adjustment}
State-of-the-art methods~\cite{pavlakos2018learning, kanazawa2018end, omran2018neural} for SMPL model estimation employ a weak-perspective camera model to facilitate the computation. 
However, as it is known, most of videos are captured with a perspective camera. So in order to better utilize the photometric loss, we adjust the camera model from the weak-perspective model to the perspective model. 
As the focal length of videos in the wild are normally unknown, we use a medium focal length to render SMPL model, and we empirically found that this perspective camera model can align SMPL model to the image well.
We take a simple conversion from the weak perspective model to a general perspective model, assuming known camera focal length as the following,
\begin{equation}
V_{tras.} = [t_x, t_y, \frac{f_c}{\frac{1}{2}*img\_size * s}],
\label{equ:translation}
\end{equation}
where vector $t=[t_x,t_y]$ is translation and $s$ is the scale in a weak-perspective camera model.  $f_c$ is the focal length of the camera.  


\begin{figure}[t]
\begin{center}
\includegraphics[width=1.0\linewidth]{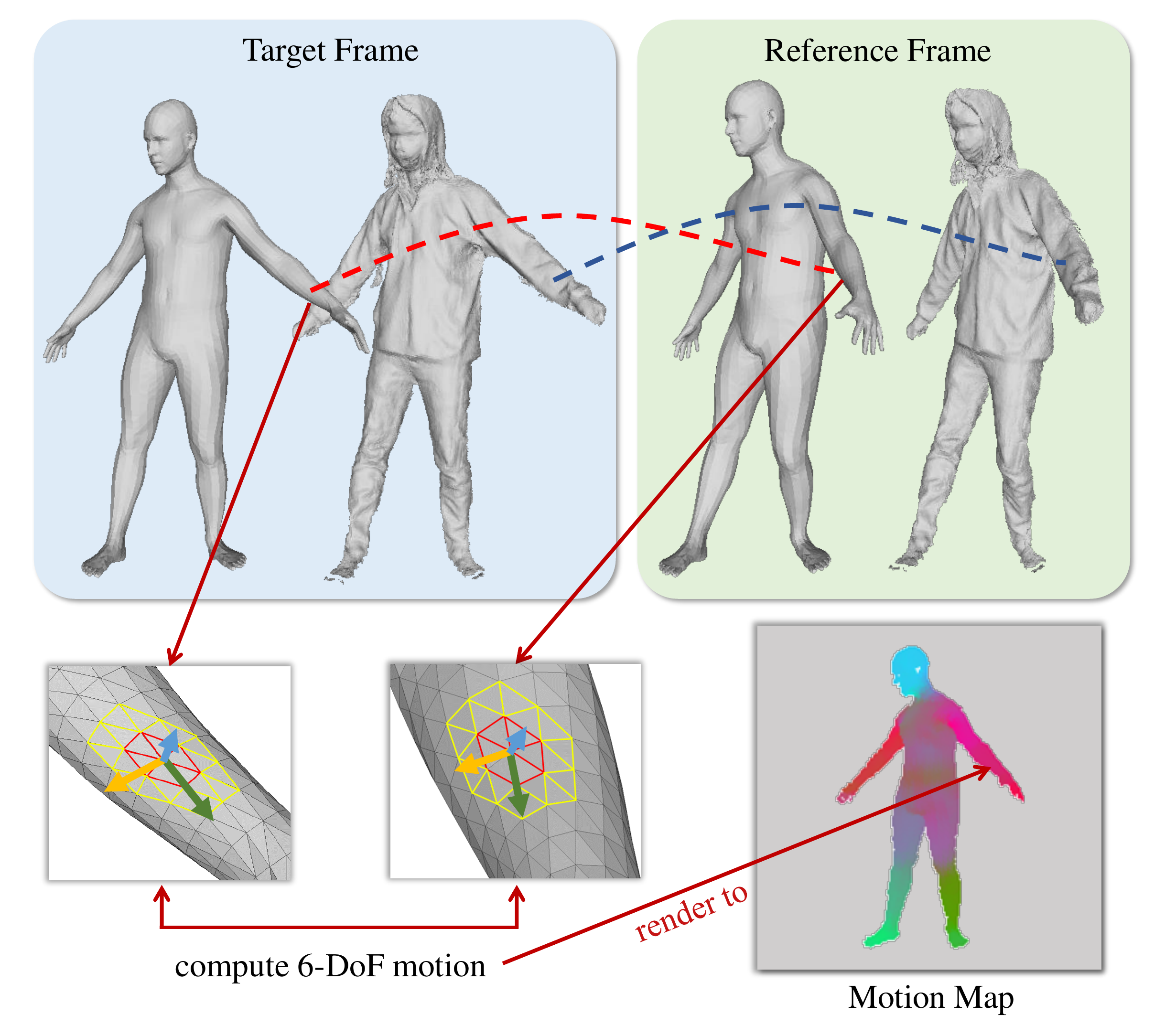}
\vspace{-0.3in}
\end{center}
   \caption{Motion map generation.  Given the SMPL models at two neighboring frames, we compute a per-vertex transformation using some nearby vertices (colored in the zoom-in figure). This per-vertex transformation is then rendered to image plane as a  motion map.  Our method assumes that the non-rigid transformation between the SMPL shapes (red dashed line) is the same as that between the detailed shapes (blue dashed line).}
\label{fig:motion_map}
\vspace{-0.2in}
\end{figure}

\subsection{Non-rigid Motion Model}\label{sec:NRMM}
In order to exploit the photometric consistency between different frames for the self-supervision, we need to compute the motion fields of the human body between consecutive frames. However, this is non-trivial as the human motion is non-rigid. 
So we propose a novel approach to compute the non-rigid motion of the human body between consecutive frames based on the estimated SMPL model per video frame.  
Specifically speaking, for each video, we select the target frame intervally with a gap of 3, then group the consecutive $r=[\pm4, \pm5, \pm7, \pm8, \pm9]$ frames as reference frames. After grouping each image tuple, we compute a non-rigid motion field $T_{t \rightarrow r}$ to represent the 3D dense spatial transformation of the human body from the target frame to the reference frame. 
The non-rigid motion is defined as a motion field in 3D space, with a 6-DoF transformation for each vertex of the SMPL model. 

To compute the non-rigid motion, we first compute a per-vertex transformation. As shown in Figure~\ref{fig:motion_map}, since the SMPL models at two neighboring frames share the same topology, we have explicit per-vertex correspondence.
Thus, the per-vertex transformation can be computed by registering $n$ two-ring neighboring vertices in the target and reference model.  More specifically, the rotation matrix $R \in SO(3)$ and translation vector $t \in \mathbb{R}^3$ can be computed by
\begin{equation}
\vspace{-0.1in}
    \begin{aligned}
        R &= \argmin \sum_{i = 1}^{n} ||R(v_t^i-{v_t^c}) -  (v_r^i-v^c_r)||^2,  \\
        t &= v_r-R*v_t,
    \end{aligned}
\end{equation}
where ${v_r^i}$ and ${v_t^i}$ represent the $i$th corresponding vertices in these two-ring neighboring vertices in reference model $r$ and target model $t$ respectively,
$v^c$ denotes the center vertex of the two-ring neighbor group.

Then we render the per-vertex transformation to the image plane as a motion map by ray tracing. For each pixel in 2D image, the mean $R$ and $t$ of the 3 vertices in the corresponding triangle is computed as the final transformation. 
In addition to motion map, we also render the depth of SMPL model in the target frame as base depth $D_t$.


\subsubsection{Occlusion Handling and Baseline Filtering}\label{validation_map}
Exploiting the motion of human shape to measure photo-consistency between neighboring frames faces two technical challenges, namely occlusion and insufficient baseline length.
When the motion is too large, part of the body might become invisible in the reference or target view, just like the occlusion problem in wide baseline stereo matching. 
When the motion is too small, adjusting shape details by maximizing photo-consistency could lead to noisy results. This is similar to the problem when the stereo baseline is too short.

To make the self-supervised training of ReconNet robust, we design careful filtering to deal with this problem. We define a `baseline length' for each pixel in the motion map, which is the magnitude of its translation $t$.
We then compute the mean baseline over all pixels for each tuple, and remove tuples with mean baseline less than $0.5$m.
To deal with the occlusion due to large motion, for each $T_{t\rightarrow r}$, we compute a validation mask $M_{r}$, where we mark a pixel as valid if it is visible in both target and reference view and has a baseline length larger than 5 cm.

\subsection{ReconNet Architecture}\label{reconnet}
The ReconNet computes the detail depth layer which will be added to the base depth to compose the final result. We adopt a variant of U-Net~\cite{ronneberger2015u} using the residual blocks~\cite{He_2016_CVPR} in encoder and decoder with skip connections. The encoder has 6 down-sampling layers, while the decoder has 5 up-sampling layers. We apply a sigmoid function at the end of the last layer to regularize the output from -10 to 10 cm. The input is a concatenation of the 512$\times$512 RGB image and the zero-median rendered depth map from the estimated SMPL model. The output of the network is a 256$\times$256 depth offset map. Finally, we compose the detail depth layer with the base depth layer to obtain our final output.


\subsubsection{Self-supervised Learning for Detail}\label{self-train}
Warping-based view synthesis loss has been proved effective in monocular, stereo, and multi-view stereo depth prediction tasks~\cite{godard2017unsupervised, zhou2017unsupervised, khot2019learning}. We extend it to monocular non-rigid human depth estimation. 

\begin{figure}
\center
\includegraphics[width =1.0 \linewidth]{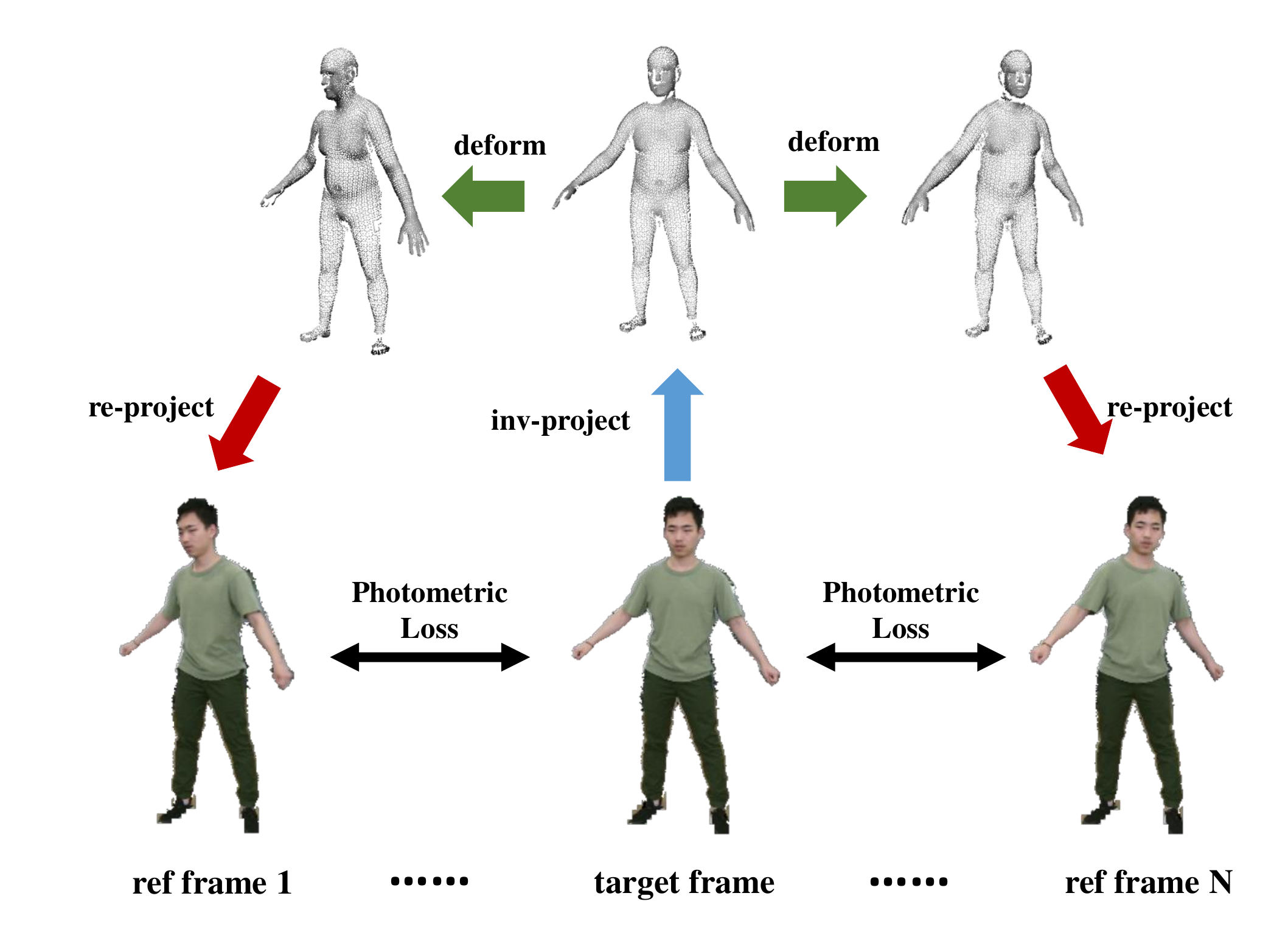}
\caption{We first inverse-project the composed depth to point clouds, then deform them with non-rigid motion map and finally reproject deformed point clouds back to reference image for bilinear sampling.
}
\label{fig:photo_loss}
\end{figure}

Given a clip of temporal continuous frames $\{I_1,...,I_N\}$ with fixed camera intrinsic parameters $K$,  we select the center frame $\{I_t\}$ as the target frame and the others as reference frames $\{I_r\}(1\leq r\leq N, r \ne t)$. For each reference frame, a non-rigid motion $\{T_{t\rightarrow r}\}_{r=1}^{N-1}$ and the validation map $\{M_{t\rightarrow r}\}_{r=1}^{N-1}$ can be pre-computed with the estimated SMPL models. Then our network use these non-rigid motion fields to warp the target frame toward the reference frames with a differentiable bilinear interpolation. This process is illustrated in Figure~\ref{fig:photo_loss}. 

Let $p_t$ denote the homogeneous coordinates of a pixel in the target view, and $K$ denote the camera intrinsic matrix. We can obtain $p_t$’s projected coordinates onto the reference view $p_r$ by
\begin{equation}
p_r \sim KT_{t \rightarrow r}(p_t)D(p_t)K^{-1}p_t.
\label{equ:warping}
\end{equation}

The inverse-warped images $\{\hat{I}_{t}^{r}\}$ from each reference frame can be synthesized according to Equation~\ref{equ:warping}. As a result, we can then formulate a photo-consistency objective function as the following:
\begin{equation}
\small
L_{photo}^r= \left( \alpha \frac{1-SSIM_{cs}(I_t, \hat{I}_t^r)}{2}
+ (1-\alpha) \parallel I_t - \hat{I}_t^r\} \parallel \right)  \otimes M_r, 
\label{equ:photo_loss}
\end{equation}
where SSIM$_{cs}$ denotes the structural similarity index~\cite{wang2004image} with only the component of contrast and structure ($SSIM_{cs}=\frac{\sigma_{xy} + c}{\sigma_x^2+\sigma_y^2+c}$). We set $\alpha$ to 0.9, because the estimated SMPL is imperfect, which causes misalignment during image warping with the computed non-rigid motion. So the structural similarity measured by SSIM$_{cs}$ is more robust than the intensity difference. 
Moreover, we also use the validation map to mask out invalid pixels. 

Our final photo-consistency function is summed over all reference images for better robustness,  
\begin{equation}
L_{photo}= \sum_{r=1,r\ne t}^{N}L_{photo}^r.
\label{equ:photo_loss_all}
\end{equation}

Following previous self-supervised depth estimation, a smoothness term is also introduced. Since our target is to estimate a human shape, we require the gradient of the composed shape to be close to the base shape, which leads to the following smooth term:
\begin{equation}
L_{smooth}= \sum_{p_t}|\triangledown D_{detail}(p_t)-\triangledown D_{base}(p_t)|.
\label{equ:smooth_loss}
\end{equation}
We also introduce a regularization term to require the final depth to be similar to the base depth: 
\begin{equation}
L_{regularizer}= \sum_{p_t}|D_{detail}(p_t)- D_{base}(p_t)|.
\label{equ:regular_loss}
\end{equation}
Finally, our final learning objective function is:
\begin{equation}
L= L_{photo} + \gamma_{s} L_{smooth} + \gamma_{r} L_{regularization},
\label{equ:total_loss}
\end{equation}
where $\gamma_{s}$ and $\gamma_{r}$ are the hyperparameters to control the significance of the smooth term $L_{smooth}$ and regularization term $L_{regularizer}$. In all our experiments, $\gamma_{s}$ is set as $10^{-5}$ and  $\gamma_{r}$ is  $10^{-6}$.

\section{Experiment}

\begin{figure*}[t]
\begin{center}
\includegraphics[width=1.0\linewidth]{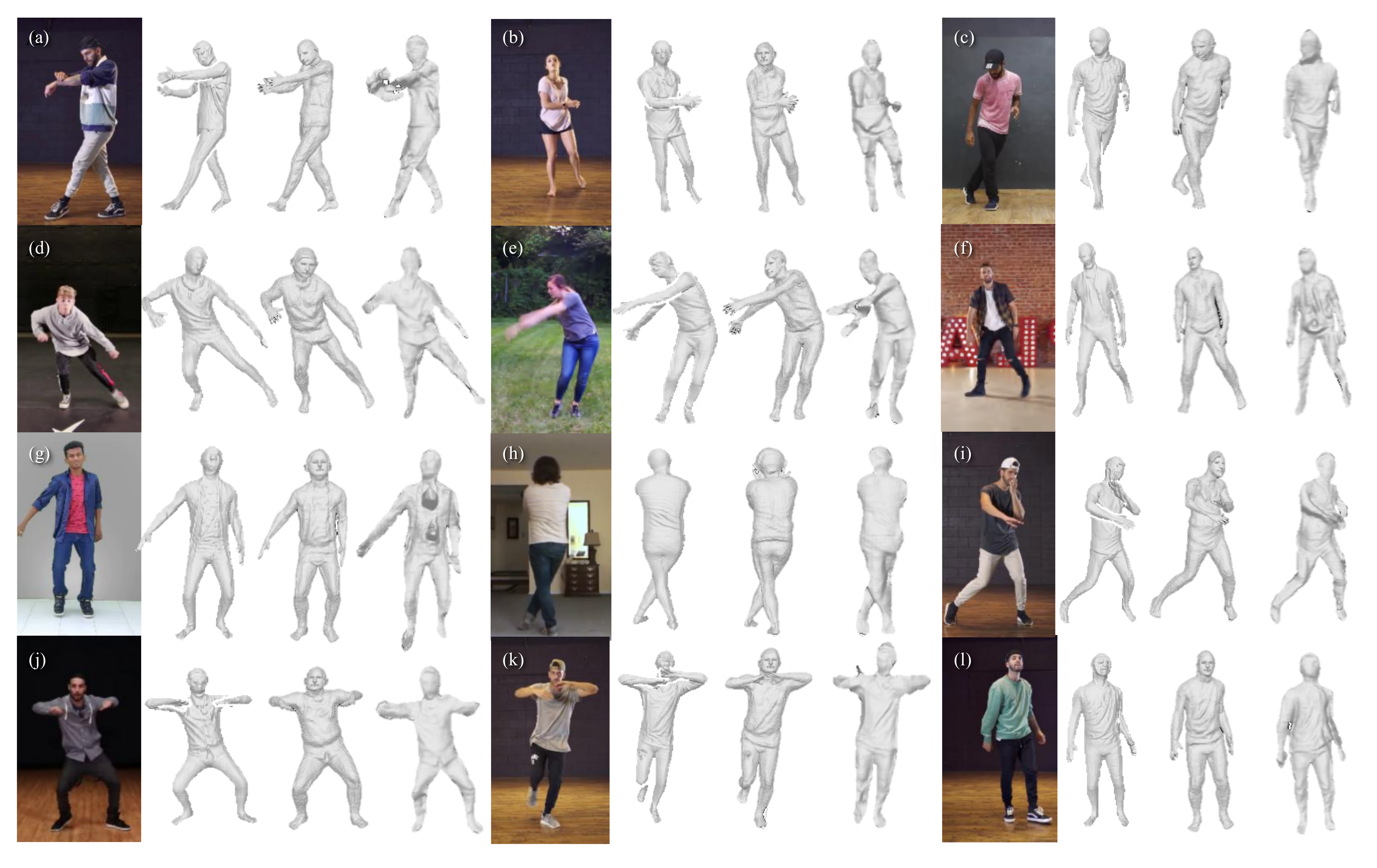}
\end{center}
    \vspace{-0.4in}
    \caption{Experiments on data in the wild. From left to right, each example shows a single input image, our result, the result from HMD\cite{zhu2019detailed}, and the result from Tang~\etal\cite{tang2019neural}.}
    \vspace{-0.1in}
\label{fig:wild_data}
\end{figure*}

\begin{figure}[t]
\begin{center}
\includegraphics[width=1.0\linewidth]{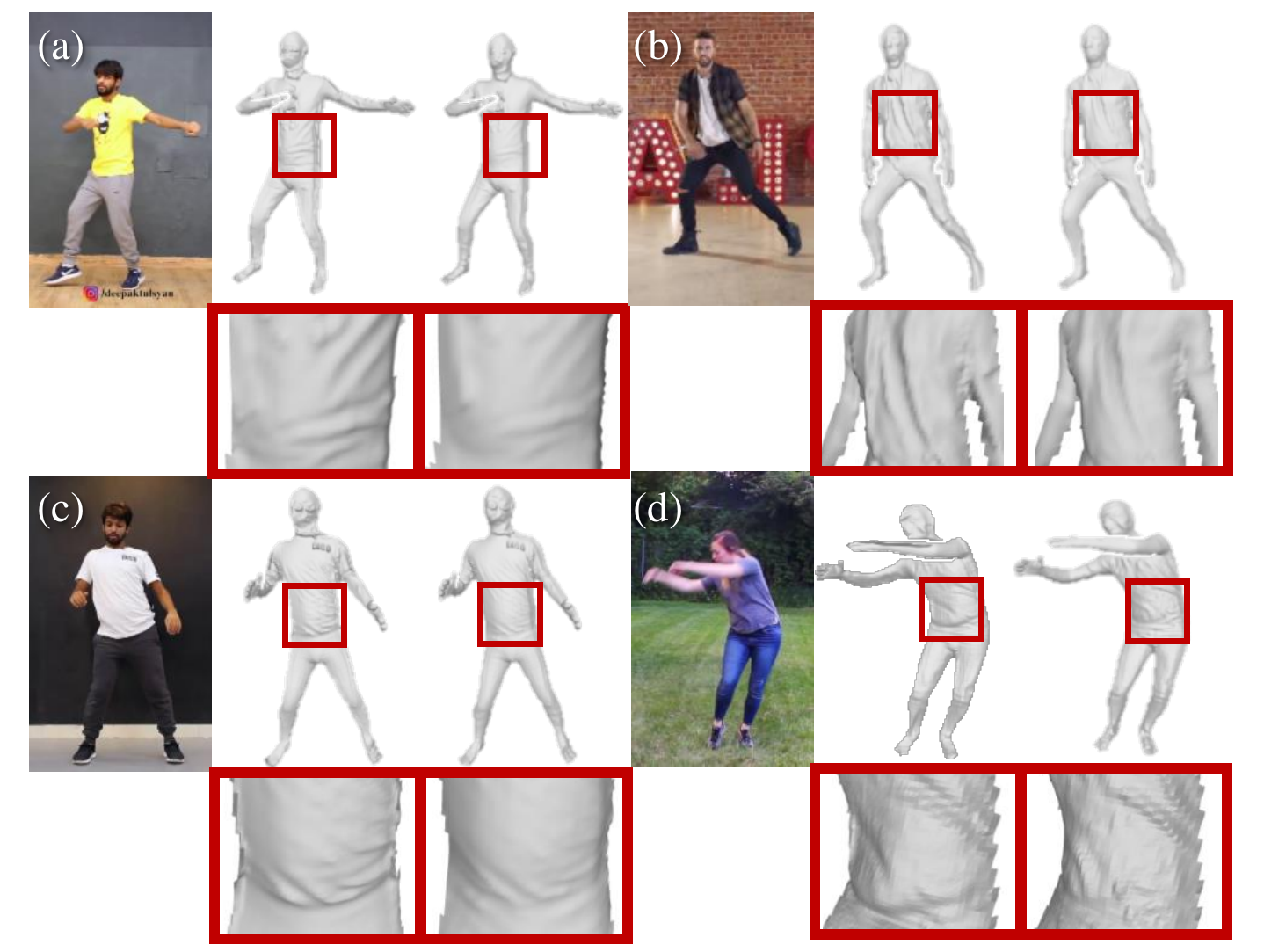}
\end{center}
\vspace{-0.2in}
   \caption{From the left to right, each example shows a single input image, the result after finetuned on YouTube data, the results before finetuning. This finetuning improves both mid- and high- frequency shape details.}
\label{fig:finetune_compare}
\vspace{-0.1in}
\end{figure}

\subsection{Data}\label{sec:data}
We find the poses tracked by original HMR model trained in the wild images are not accurate enough for the self-supervised learning of ReconNet, so we finetune the TrackNet using our collected data with ground truth SMPL parameters generated by DoubleFusion\cite{yu2018doublefusion}.  

Thus, we captured 36 video sequences of different people performing simple action, which contains roughly 48,000 frames in total.  Half of the frames have 
labeled SMPL coefficients recovered from depth streams by DoubleFusion~\cite{yu2018doublefusion}, 
and
are used in training of the TrackNet. To augment the background in the video, we randomly use images from the Places Dataset\cite{zhou2017places} as the background for each sequence.  All the image frames are also used to bootstrap our ReconNet in a self-supervised manner, and the SMPL parameters are predicted by our finetuned TrackNet. To train the ReconNet, we grouped the frames from the captured videos. For each clip, we skip every 3 frames to set a target frame, and other consecutive $[\pm4, \pm5, \pm7, \pm8, \pm9]$ frames are set as the reference frames. We sampled in total 12,533 image tuples for training. 

To ensure our model can be generalized to in-the-wild data, we select 18 videos from YouTube, and generate about 3,000 images tuples to finetune our ReconNet. We select YouTube videos with a simple criteria that the video contains a \emph{single} and \emph{complete} person with less occlusion.  Note that we can replace TrackNet with other better SMPL model or even SMPL-X \cite{SMPL-X:2019} model estimation networks for more accurate base shape and motion map generation.


\subsection{Training Details}\label{sec:train}
We finetune the TrackNet from the original HMR model with `Adam' optimizer using our captured data with SMPL parameters from DoubleFusion.  The learning rate is set to $1\times10^{-6}$.  We use batch size $20$ and train in $20$ epochs.


We first bootstrap the ReconNet with our captured videos in self-supervised manner for 2 epochs. The learning rate is set as $4\times10^{-4}$ and the batchsize is 2. We then finetune the ReconNet with YouTube images with the learning rate of $1\times10^{-4}$ for one epoch.

\begin{figure*}[t]
\begin{center}
\includegraphics[width=1.0\linewidth]{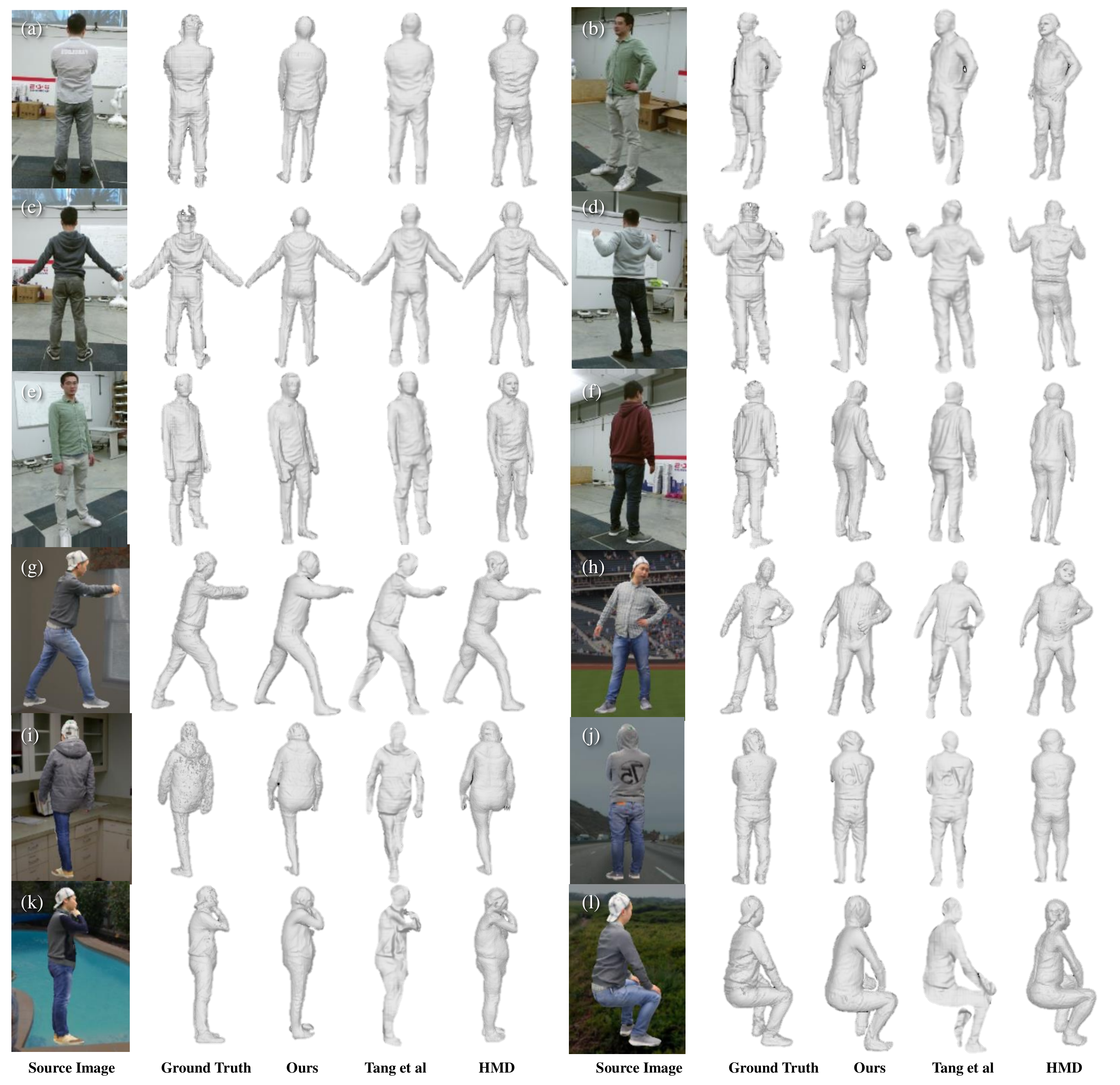}
\end{center}
    \vspace{-0.2in}
    \caption{Comparison to Tang~\etal~\cite{tang2019neural} and HMD\cite{zhu2019detailed} on their testing datasets.  The source images in top half part are from the testing data of Tang\etal~\cite{tang2019neural}, and the source images in bottom half part are from the REAL dataset in HMD\cite{zhu2019detailed}.}
    \vspace{-0.12in}
\label{fig:compare_tang}
\end{figure*}

\subsection{Experiment on Data in the Wild}
We test our method on unseen YouTube videos. We randomly select half of the frames in the video to finetune our ReconNet, and use the other frames for evaluation. Figure~\ref{fig:wild_data} shows the comparison with HMD\cite{zhu2019detailed} and Tang~\etal~\cite{tang2019neural} on in-the-wild data. We find our result can capture more fine details compared with other two methods mainly for two reasons: first, our model is finetuned with in-the-wild videos due to the self-supervised learning; second,  our photometric loss is more effective in capturing small wrinkles. In comparison, the other two methods are trained only with limited `ground truth' depth from consumer depth cameras. The noisy `ground truth' depth makes it difficult for the network to recover small details. Further more, although both our method and Tang~\etal~\cite{tang2019neural} separate the human shape to a base shape and a detail shape, we use a SMPL model for the base shape, which is more robust for the in-the-wild data. It is clear that our method generates more details than \cite{tang2019neural} and \cite{zhu2019detailed} in the examples (a), (b), (d), (h), (j). Tang~\etal's\cite{tang2019neural} results have large errors on examples (a), (b), (e), (g), (k). 
We notice that the complex clothing, faces and hairstyles are not estimated accurately, which is mainly because they are difficult for photo-consistency based reconstruction. 

We also show the predicted results with and without the fine-tuning the ReconNet on Youtube data in Figure~\ref{fig:finetune_compare}. Our model cannot capture the fine details without finetuning. Finetuning on YouTube data improves the mid- and high- frequency shape details. 
Please refer to the supplementary material for more results and discussion.

\subsection{Quantitative Evaluation}
To quantitatively evaluate the accuracy and compare to the previous methods, we evaluate our method on the testing data provided by Tang~\etal\cite{tang2019neural} and HMD~\cite{zhu2019detailed}.

\textbf{Comparison on the Dataset from~\cite{tang2019neural}.} 
Tang~\etal \cite{tang2019neural} published a small dataset with ground truth 3D human shapes generated by InfiniTAM~\cite{InfiniTAM_ISMAR_2015} for quantitative evaluation.
To evaluate our results on their dataset, we first use ICP to register our results to the ground truth, and measure the error at each pixel by the point-to-point nearest neighbor distance.
Following~\cite{tang2019neural}, we compute the accuracy at different error thresholds, i.e. the percentage of pixels with an error smaller than some threshold.
The accuracy of different methods evaluated on this dataset is shown in Table~\ref{tab:tang}. We also compute the Mean Absolute Error (MAE) to evaluate the overall shape accuracy.
Our method has smaller MAE both than Tang~\etal~\cite{tang2019neural} and HMD, and higher accuracy when the error threshold is larger than 4cm. It suggests our method has less large errors than~\cite{tang2019neural} on their published dataset. However, Tang~\etal  can recover better shape details on this dataset. We believe this is because the test data is highly consistent with their training data. The advantages of our self-supervised method is demonstrated on data in the wild in Figure~\ref{fig:wild_data}. The performance of HMD~\cite{zhu2019detailed} varies on this dataset, e.g. poor results on example (b), (d) and (f), suggesting its generalization is poorer than our method.

The first three rows in Figure~\ref{fig:compare_tang} shows some visual results from these methods. Tang~\etal's method sometimes generate distorted limbs as shown in example (b) and (d).

\textbf{Comparing on the Dataset from~\cite{zhu2019detailed}.}  
HMD provides another small dataset with ground-truth 3D mesh recovered by multi-view reconstruction methods for quantitative evaluation. Here, we focus on comparing our ReconNet with the `Shading-Net' in HMD, which is also a refinement model on an estimated SMPL model. For fair comparison, we use the same SMPL model for both methods. The accuracy is reported in Table~\ref{tab:hmd_test}.  We can see our self-supervised method achieves better accuracy than HMD even on their test data. 

The last three rows in Figure~\ref{fig:compare_tang} are the qualitative comparison between these three methods on this dataset. We can find Tang~\etal's performance in this dataset is not good because of its poor generalization and the unusual poses in this dataset. Also, we can find our method can recover more details in example (g), (j), (k).

\subsection{Ablation Study}
We perform various ablation studies on the dataset of Tang~\etal\cite{tang2019neural} in this subsection.
We denote the result without using validation mask $M_{r}$ and SSIM$_{cs}$ as baseline, and compare it with various other settings. All the ablation study are trained with the same hyperparameter. 

\textbf{Validation Mask} 
The baseline method generate poorer accuracy than the proposed method with validation mask as shown in Table~\ref{tab:ablation_iccv}. This proves the effectiveness of our occlusion handling and baseline filtering. 


\textbf{SSIM$_{cs}$ loss.} We replace the original SSIM loss with SSIM$_{cs}$ loss in training the ReconNet. SSIM$_{cs}$  is measured on the contrast domain and is more robust to misalignment (due to imperfect non-rigid motion estimation) and shading changes. 
As shown in Table~\ref{tab:ablation_iccv}, after replacing SSIM with SSIM$_{cs}$, our model achieves better performance.




\begin{table}[]
\caption{Comparison on the dataset published in \cite{tang2019neural}.}
\vspace{-0.1in}
\begin{tabular}{|l|lll|l|}
\hline
\multicolumn{1}{|c|}{Methods} & \multicolumn{3}{|c|}{Accuracy}                                                                           & MAE \\ \cline{2-4}
\multicolumn{1}{|c|}{} & \multicolumn{1}{c|}{1.0cm} & \multicolumn{1}{c|}{2.0cm}& \multicolumn{1}{c|}{4.0cm} &                      \\ \hline
Tang~\etal\cite{tang2019neural} & \multicolumn{1}{l|}{\textbf{33.30}} & \multicolumn{1}{l|}{\textbf{59.68}} &  79.63 & 2.735\\
\hline
HMD~\cite{zhu2019detailed}  & \multicolumn{1}{l|}{27.66} & \multicolumn{1}{l|}{54.10} & 76.31             & 3.077
\\
\hline
Ours & \multicolumn{1}{l|}{31.47} & \multicolumn{1}{l|}{59.08} & \textbf{82.13}             & \textbf{2.609}       \\ \hline
\end{tabular}
\label{tab:tang}
\vspace{-0.1in}
\end{table}

\begin{table}[]
\caption{Comparison on the dataset published in \cite{zhu2019detailed}. `Ours (HMD)' means our method fed with the same undressed SMPL model as HMD.}
\vspace{-0.1in}
\begin{tabular}{|l|l|l|l|l|}
\hline
\multicolumn{1}{|c|}{\multirow{2}{*}{Methods}} & \multicolumn{3}{c|}{Accuracy (\%)}               & \multicolumn{1}{c|}{\multirow{2}{*}{\begin{tabular}[c]{@{}c@{}}MAE\\ (cm)\end{tabular}}} \\ \cline{2-4}
\multicolumn{1}{|c|}{}         & 1.0cm     & 2.0cm    & 4.0cm  & \multicolumn{1}{c|}{}     \\ \hline
Tang~\etal\cite{tang2019neural}  & 19.07 & 41.71 & 73.19  & 3.125            \\ \hline
HMD~\cite{zhu2019detailed}  & 21.83 & 46.10 & 75.46  & 3.043            \\ \hline
Ours(HMD)       & \textbf{22.62} & \textbf{47.65}   & \textbf{76.65} & \textbf{2.944}    \\ \hline
\end{tabular}
\label{tab:hmd_test}
\vspace{-0.1in}
\end{table}

\begin{table}[]
\caption{Ablation study on Tang~\etal's test set. Please see text for more details.
}
\vspace{-0.1in}
\begin{tabular}{|l|lll|l|}
\hline
\multicolumn{1}{|c|}{Methods} & \multicolumn{3}{|c|}{Accuracy}                                                                           & MAE \\ \cline{2-4}
\multicolumn{1}{|c|}{}                        & \multicolumn{1}{c}{1.0cm}           & \multicolumn{1}{c}{2.0cm}           & \multicolumn{1}{c|}{4.0cm} &                      
\\ \hline


Ours(Baseline)       &   \multicolumn{1}{l|}{28.14}          & \multicolumn{1}{l|}{55.57}          & 79.46                      & 2.828                \\ \hline
Ours(M)                           & \multicolumn{1}{l|}{28.52}          & \multicolumn{1}{l|}{56.35}          & 80.67                      & 2.714                \\ \hline
Ours(M+SSIM$_{cs}$)                             & \multicolumn{1}{l|}{\textbf{31.47}} & \multicolumn{1}{l|}{\textbf{59.08}} & 
\textbf{82.13} & \textbf{2.609}       \\ \hline

\end{tabular}
\label{tab:ablation_iccv}
\vspace{-0.2in}
\end{table}

\section{Conclusions}\label{sec:conclusion}
We present a self-supervised method to estimate human shapes with fine geometry details such as cloth wrinkles. This self-supervised approach enables the network to be trained on in-the-wild data, such as YouTube videos, which significantly improves the generalization of the  network. This result is achieved by introducing the SMPL model for the base shape, and using it to compute the non-rigid human motion at neighboring frames to facilitate photo-consistency evaluation. Extensive evaluation and comparison with state-of-the-art methods proves the effectiveness of our method.


{\small
\bibliographystyle{ieee_fullname}
\bibliography{egbib}
}

\end{document}